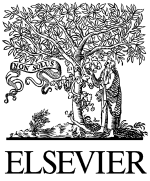

ELSEVIER

# Gradient Difference Based Approach for Text Localization in Compressed Domain

B.H.Shekar[a,*] , Smitha M.L.[b,*]

[a]*Department of Computer Science, Mangalore University, Mangalore, Karnataka, India.*
[b]*Department of Master of Computer Applications, KVG College of Engineering, Sullia, Karnataka, India.*

**Abstract**

In this paper, we propose a gradient difference based approach to text localization in videos and scene images. The input video frame/ image is first compressed using multilevel 2-D wavelet transform. The edge information of the reconstructed image is found which is further used for finding the maximum gradient difference between the pixels and then the boundaries of the detected text blocks are computed using zero crossing technique. We perform logical AND operation of the text blocks obtained by gradient difference and the zero crossing technique followed by connected component analysis to eliminate the false positives. Finally, the morphological dilation operation is employed on the detected text blocks for scene text localization. The experimental results obtained on publicly available standard datasets illustrate that the proposed method can detect and localize the texts of various sizes, fonts and colors.

## 1. Introduction

Text localization in document/scene images and video frames aims at designing an advanced optical character recognition (OCR) systems. However, the large variations in text fonts, colors, styles, and sizes, as well as the low contrast between the text and the complicated background, often make text detection extremely challenging. The researcher's experimental results on such complex text images/video reveal that the application of conventional OCR technology leads to poor recognition rates. Therefore, efficient detection and segmentation of text blocks from the background is necessary to fill the gap between image/video documents and the input of a standard OCR system.

The text-based search technology acts as a key component in the development of advanced image/video annotation and retrieval systems. Detection of text from color video images has unique problems. Video frames are typically low contrast, multi-colored and compressed which can cause color bleeding between the text and the background. With the wide use of computers and large scale storage and transmission of data, efficient ways of storing of video data have become necessary. Dealing with such enormous information can often present difficulties in the real scenario. To overcome this problem, researchers suggest image compression technique which minimizes the size in bytes of a graphics file without degrading the quality of the image to an acceptable level. The reduction in file size allows more images to be stored in a given amount of disk or memory space. Therefore, image compression is achieved by removing data redundancy while preserving information content.

In this paper, we introduce a gradient difference based approach in compressed domain which helps in detecting and localizing the text effectively. Certain heuristic rules are employed on the gradient information where the zero crossings are identified to fix the bounding boxes and hence the text is localized. The remaining part of the paper is organized as follows. The related works are presented in section 2. The proposed approach is discussed in section 3. Experimental results and comparison with other approaches are presented in section 4 and conclusion is given in section 5.

\* Corresponding author. Tel.:+919480146921, +919945427570
  *E-mail address*: bhshekar@gmail.com, smithaml.urubail@gmail.com.

## 2. Related Works

Several methods for text extraction from video have been proposed in the last decade, which are classified as connected component based (CC), texture based, and edge and gradient based methods. A brief review on these approaches and their limitations are presented below.

Shivakumara *et al.* [20] proposed an edge based technique for text detection in images works well if the text is in horizontal direction. The frame was segmented into 16 non-overlapping blocks. The median filter and edge analysis was used to identify the candidate text blocks. Using block growing method, the complete text block was obtained. Finally, the true text regions are detected based on the vertical and horizontal projection profile analysis. In [22], filters and edge analysis were used for initial text detection. The straightness and cursiveness edge features were used for false positive elimination. A hybrid system for text detection based on the edges, local binary pattern operator, and SVM was proposed by Anthimopoulos *et al.* [16, 17]. Text detection using a cascade AdaBoost classifier with HOG and multi-scale local binary pattern feature was proposed by Pan *et al.* [39]. Text localization was done using window grouping technique. Within each located text line, local binarization is done to extract candidate CCs and non-text CC's are filtered using Markov Random field model and MLP in order to get the final text line.

Gradient difference based method for text detection was proposed by Shivakumara *et al.* [21]. Zero crossing was used to determine the bounding boxes for the detected text line. Moradi *et al.* [18] used edge and corner detection method and discarded non-text corners by histogram analysis. A Self-Organizing Map (SOM) neural network based technique for artificial text detection in video frames was due to Yu *et al.* [7]. Three layers of supervised SOM were used to classify text, and non-text areas. Huang *et al.* [36] used the texture feature in the stroke map to detect text. For text localization, Harris corner detection approach was used on the stroke map. Morphological operations were used to connect the corners. Guru *et al.* [4] proposed an eigen value based technique, which performs a block wise eigen analysis on the gradient image of the video frame. Eigen analysis helped in identifying the potential text blocks. Shivakumara *et al*. [23] introduced the classification of low and high contract images for text detection. They analyzed the number of edges found using sobel and canny edge detector for low and high contrast images, to form the heuristic rules for classification.

Wavelet transforms and its variants have become very popular among researchers for texture analysis. Most of the recent works on texture based text detection and localization are based on the wavelet transform [19, 24, 27, 29, 41]. Other methods such as the Gabor filter [1, 38], DCT [38], Haar wavelet [31], Spatial analysis [16], Laplacian [28], Fourier [26] etc. were also used by researchers in the recent past. Ye *et al*. [29] used 2D wavelet coefficients to calculate histogram wavelet coefficients of all pixels. The SVM with a RBF kernel was used for classification of text and non-text. They introduced an OCR feedback procedure to locate the final text lines. Zhao *et al*. [19] used wavelet transform and sparse representation with discriminative dictionaries for text detection. Shivakumara *et al*. [25, 27] also used Haar wavelet in both the works. In [25], *k-means* clustering was used to classify text and background. In [27], they also used color features along with Wavelet-Laplacian method to detect text. Discrete Cosine transform (DCT) was used by Qian *et al*. [38]. Texture intensities were used to verify horizontal and vertical text. Horizontal and Vertical projection profiles were used for text localization. Yi *et al*. [40] used Gabor filters to describe the stroke components in the text characters. They defined Stroke Gabor Words (SGWs) and used it with image window classification techniques to detect text regions. Peng *et al*. [39] computed the features using 2-D Gabor filters and Harris corner detection. Based on the confidence of text and background labeling by SVM, Conditional Random Field framework is defined and isolated text blocks are merged by heuristic reasoning. Phan *et al*. [33] used the same Laplacian approach as in [28] to identify text candidates, but used CC analysis to form simple CCs. Using the straightness and edge density feature the text blocks were finalized. Shivakumara *et al*. [25] used wavelet- median moment feature with k-means clustering to obtain text pixels. Angle projection based boundary growing was used to handle multi-oriented text.

From the literature review, it is realized that the connected-component based methods [28, 39] are simple to implement but are not robust since they assume that the text pixels belonging to the same connected region share common features such as color or grey intensity. On the other hand, texture based methods [19, 24, 27, 29, 41] may be unsuitable for small fonts and poor contrast text. In contrast to the preceding two approaches, edge and gradient based methods [18, 23] are fast and efficient but give more false positives when the background is complex. However, the major problem of these methods is in choosing threshold values to classify between text and non text pixels. To overcome these problems, the method based on uniform colors in L* a* b* space is also proposed in  [7, 8] to locate uniform colored text in video frames. Obviously, this method fails when text in video contains multiple colors in a text line or in a word. The wavelet and the SVM combination are used for text detection in the images which performs well but they include a large number of features and extensive training with the classifier [16]. The above observations lead us to devise a method to localize text with lesser false alarms.

## 3. Proposed Methodology

The video images are often degraded during transmission but the text portions can always be distinguished due to its discriminative pixel values with respect to the background. Text portions in an image always have distinct intensity values with respect to its background. The differences in the pixel values of an image are noted in the gradient of that image. Based on this observation, we propose to use gradient and edge information under compressed domain to localize the text in videos. The flow chart of the proposed methodology is illustrated in Fig. 3.1.

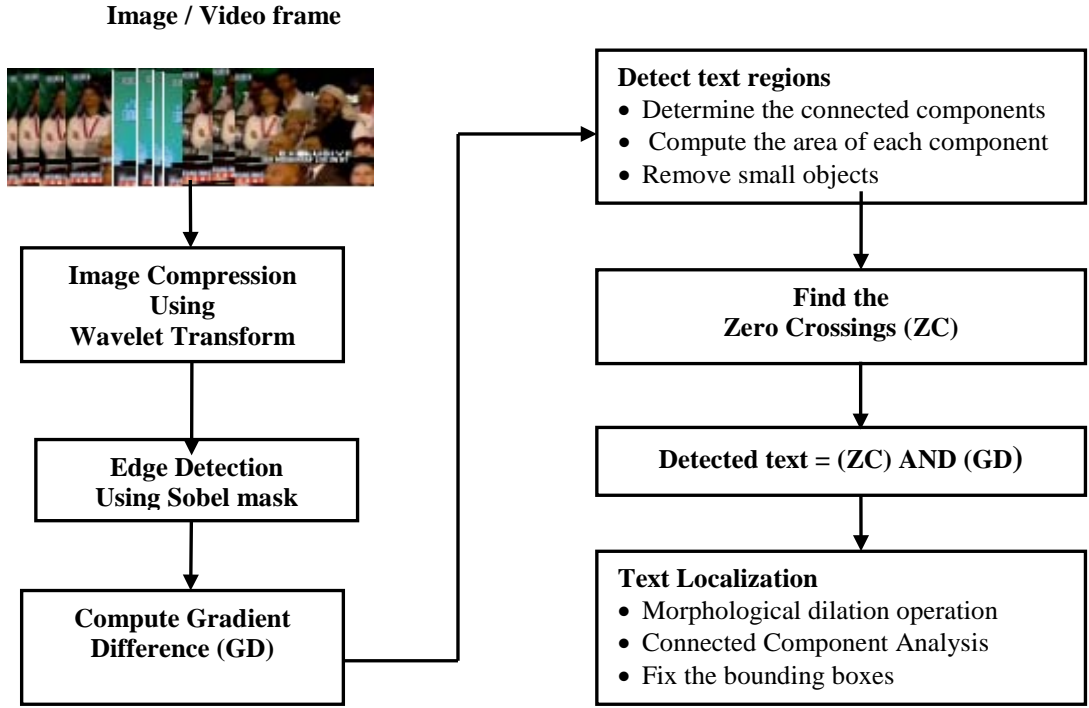

Fig. 3.1 Flowchart of the proposed method

### 3.1 Image Compression using Wavelet Transform

Image compression using wavelets is a technique used for storing a visual image that reduces the amount of digitized information needed to store the image electronically [41]. Image compression applications reduce the size of an image file without causing major degradation to the quality of the image. Image Compression using multilevel 2-D wavelet decomposition performs compression process of a signal or an image using wavelets [24, 27]. The compression procedure contains three steps: namely Wavelet Decomposition, Detail coefficient thresholding i.e. for each level from 1 to N, a threshold is selected and hard thresholding is applied to the detail coefficients and finally reconstruction to restore back the original signal.

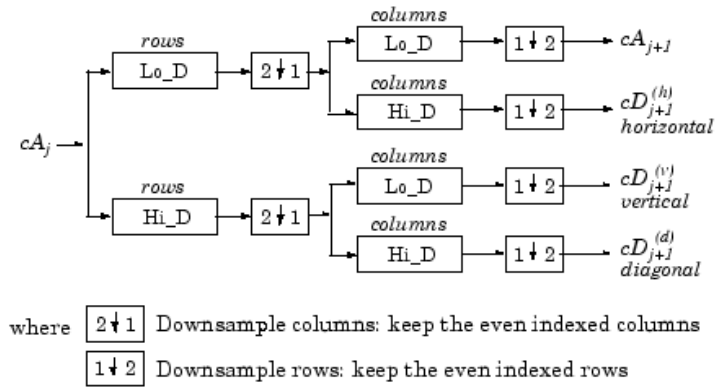

Fig. 3.2 Wavelet Decomposition Steps

- **Wavelet Decomposition**

  Digital images are very large in size and hence occupy larger storage space. Due to their larger size, they take larger bandwidth and this makes it inconvenient for storage as well as file sharing. To overcome this problem, the images are compressed in size with orthogonal wavelets called Daubechies. This compression not only helps in saving storage space but also enables a scaling function which generates an orthogonal multiresolution analysis. The two-dimensional DWT leads to a decomposition of approximation coefficients at level j in four components: the approximation at level j+1, and the details in three orientations (horizontal, vertical, and diagonal). The basic wavelet decomposition step for images is shown in Fig.3.2 where, 'L' is low-pass filter, 'H' is high-pass filter and '↓2' is down sampling.

- **Decomposition Algorithm**

  Wavelet transform (WT) represents an image as a sum of wavelet functions with different locations and scales. Any decomposition of an image into wavelets involves a pair of waveforms: one to represent the high frequencies corresponding to the detailed parts of an image (wavelet function $\psi$) and one for the low frequencies or smooth parts of an image (scaling function $\emptyset$). DWT is a multi resolution decomposition scheme for input signals. The original signals are first decomposed into two subspaces, low-frequency (low-pass) subband and high-frequency (high-pass) subband. For the classical DWT, the forward decomposition of a signal is implemented by a low-pass digital filter H and a high-pass digital filter G. Both digital filters are derived using the scaling function $\Phi$ (t) and the corresponding wavelets $\Psi$ (t). The system down samples the signal to half of the filtered results in the decomposition process. For a given signal sampling, first approximate f and fj with f, through the decomposition theorem, it is decomposed ck into fj and dk as shown below.

  Set

  $$f_j(x) = \sum_{k \in z} C_k^j \; \varphi \, (2^j \; x - k) \; \in v_j \qquad (1)$$

  $f_j$ can be broken down into $f_j = w_{j-1} + f_{j-1}$,

  in which

  $$w_{j-1} = \sum_{k \in z} d_k^{j-1} \; \psi \, (2^{j-1} \; x - k) \; \in w_{j-1} \qquad (2)$$

  $$f_{j-1} = \sum_{k \in z} C_k^{j-1} \; \varphi \, (2^{j-1} \; x - k) \; \in v_{j-1} \qquad (3)$$

- **Wavelet Reconstruction**

  In wavelet analysis, when a signal gets decomposed we need to restore it and see that if we can get the original signal. Here, we need to introduce the wavelet multistage reconstruction.

- **Reconstruction Algorithm**

  The multistage of wavelet reconstruction process worked as follows:

  Let $C_k^0, d_k^0$ and $C_i^1$ seek $C_k^1$. Then, $d_k^1$ is obtained by $C_i^2$ and finally, the $C_k^{j-1}$ and $d_k^{j-1}$ is obtained by $C_i^j$

  Set

  $$f \; = f_0 + w_0 + w_1 + \dots + w_{j-1} \qquad (4)$$

  in which,

  $$f_0 \; (x) = \sum_{k \in z} C_k^0 \; \varphi \, (x - k) \; \epsilon \, v_0 \qquad (5)$$

  $$w_n(x) \; = \sum_{k \in z} d_k^n \; \psi \, (2^n \; x - k) \; \epsilon \, w_n , \qquad (6)$$

  then,

  $$f(x) = \sum_{k \in z} C_i^j \; \varphi \, (2^j \; x - l) \; \epsilon \, v_j \qquad (7)$$

The 2-D multilevel Discrete Wavelet Transform is first applied on the input image/video frame. This produces as many coefficients as there are pixels in the image. These coefficients can then be compressed more easily because the information is statistically concentrated in just a few coefficients. This compressed image is again reconstructed using 2-D inverse discrete wavelet transform. It is observed that this reconstructed image after wavelet compression contains more fine details than the original image as reflected in Fig. 3.3 which is used for further processing.

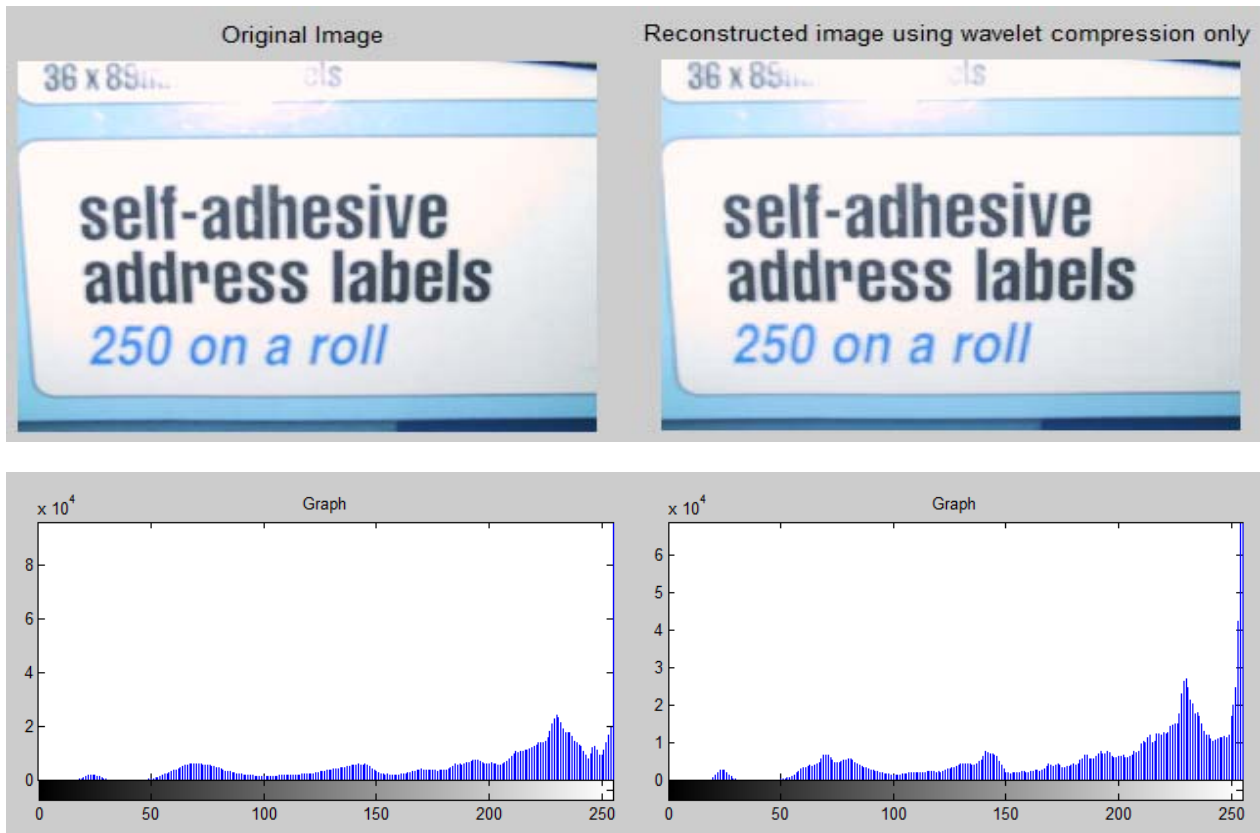

Fig. 3.3 Illustration of Image Compression using Wavelet Transform

### 3.2 Computation of Gradient Difference

The gradient information in text areas differ from non text regions because of high contrast of text [21]. We first find the edge information of the reconstructed image using sobel mask as shown in Fig. 3.4. Once the edge information is obtained, we compute the gradient *dx for* the image (G) by using a horizontal mask [-1 1]. Then, the Gradient Difference (GD) is obtained for each pixel in G as the difference between the maximum and minimum gradient values within a local window of size 1×n centered at the pixel where n is a value that depends on the character's stroke width. We have chosen n = 11 by keeping small fonts in mind. High positive and negative gradient values in text regions result from high intensity contrast between the text and background regions. Therefore, text regions will have both large positive and negative gradients in a local region due to even distribution of character strokes. This results in locally large gradient difference values as shown in Fig. 3.5 where we can see the text clearly as white patches and background as dark color.

More formally, the algorithm for detecting texts in images/video frames is as follows. Let F(x, y) be the given edge image obtained using Sobel operator and G(x, y) be the gradient image obtained by convolving horizontal mask [-1 1] with F(x,y), and W(x, y) be the local window centered at (x,y) of size 1x11. Obtain the minimum and the maximum gradient values in W over G(x,y) as follows:

$$Min\,(x\,,y) = \min_{x_i\,,\,y_i\,\in\,W(x\,,\,y)} (G(x_i,\ y_i)) \qquad (8)$$

$$Max\,(x\,,y) = \max_{x_i\,,\,y_i\,\in\,W(x\,,\,y)} (G(x_i,\ y_i)) \qquad (9)$$

Using equation (8) and (9), we compute GD(x,y) as follows:

$$GD\,(x,y) = Max\,(x,y) - Min\,(x,y) \qquad (10)$$

The resultant image obtained after computing gradient difference contains some non-text blocks which are eliminated using connected component analysis as illustrated in Fig. 3.5.

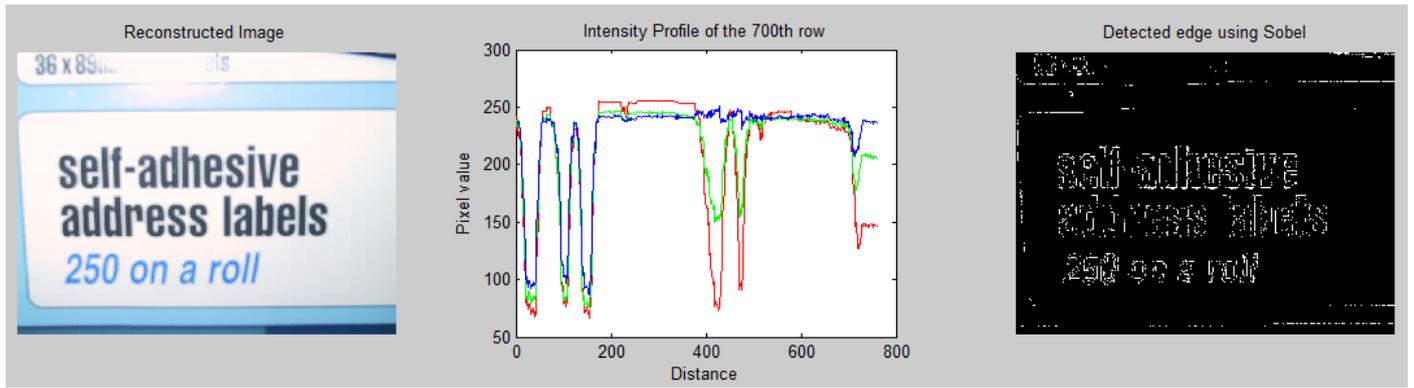

Fig. 3.4 Illustration of obtaining edge information from Reconstructed Image

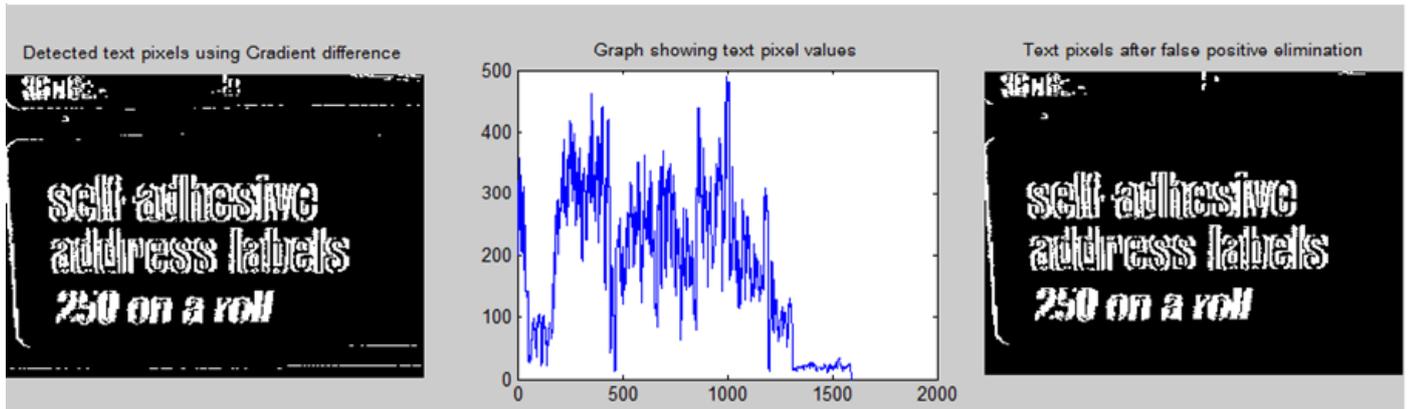

Fig. 3.5 Illustration of obtaining text clusters after finding Gradient Difference

### 3.3 Find Zero-crossings

The boundaries for white patches representing text lines are computed using a zero crossing technique [28]. Zero crossing means transition from 0 to 1 and 1 to 0. This method counts the number of transitions from 0 to 1 and 1 to 0 in each column from top to bottom of GD(x,y). We determine the connected components using 8-connectivity neighborhood of pixels and compute the area of each component. Further, the small components are eliminated by employing connected component labeling. The horizontal, vertical details and the detected zero crossings are shown in Fig. 3.6.

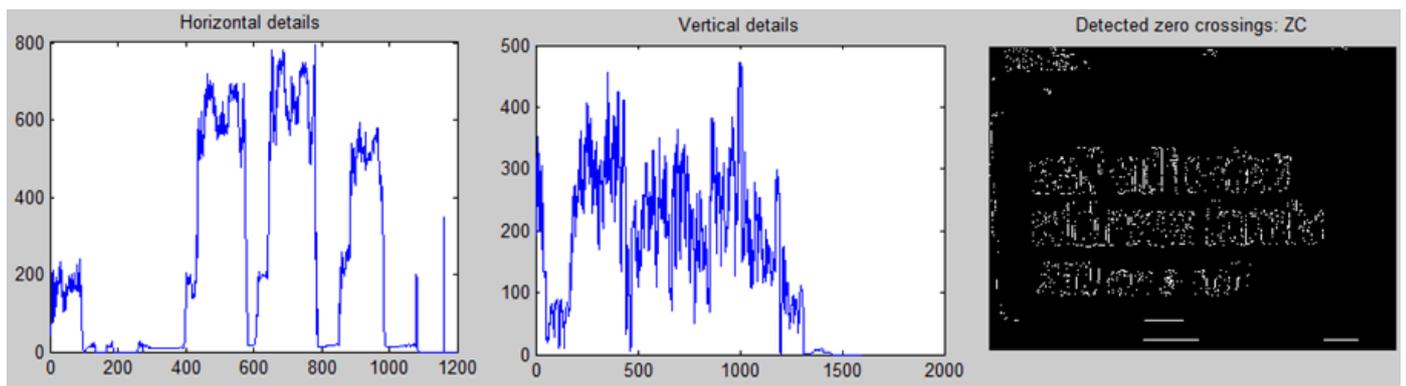

Fig. 3.6 Illustration of obtaining text clusters after identifying the zero crossings

### 3.4 Text Localization

The objective of text localization is to place rectangles of varying sizes covering the text regions [6]. We start by merging all the text detections. We perform the logical AND of the text pixels obtained by the results of Gradient Difference (GD) and the Zero Crossing (ZC) results for detecting the text. Then, morphological dilation is performed to fill the gaps inside the obtained text regions which yield better results and the boundaries of text regions are identified and finally the text is localized as illustrated in Fig. 3.7. There exists some false positives which will be addressed in our future works.

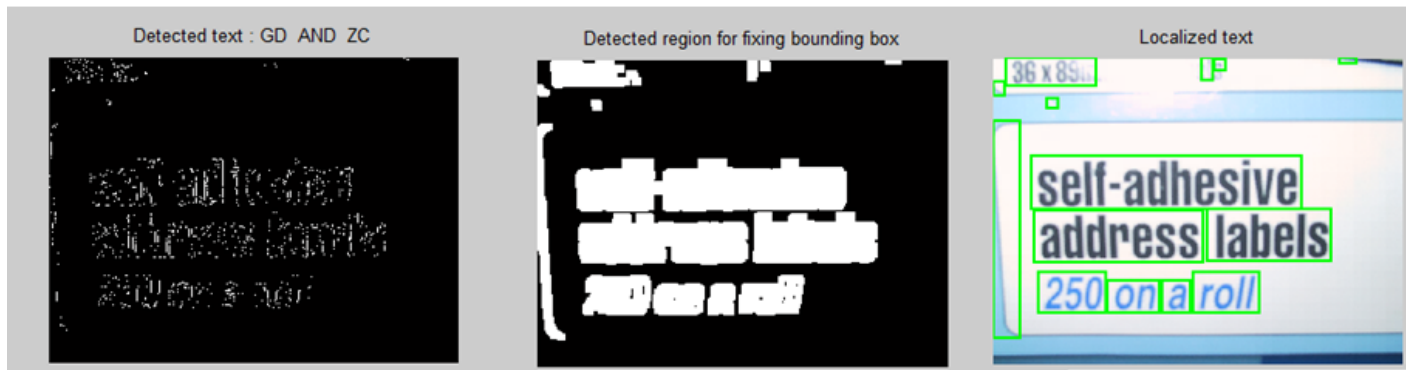

Fig. 3.7 Illustration of Text Localization

## 4. Experimental Results

This section presents the experimental results to reveal the success of the proposed approach. To estimate the performance of our system, we have conducted experimentation on ICDAR-2003, Horizontal-1 and Hua's dataset which are said to be the bench mark datasets and considered by many researchers to evaluate their approaches. We have also made a comparative analysis with some of the well known algorithms to exhibit the performance of the proposed system which is on-par with the state-of-the-art text localization approaches [3, 5, 21, 34]. The proposed approach is evaluated with respect to f-measure which is a combination of two measures: precision and recall. It is observed that most of the text blocks of each video frame possess properties such as varying fonts, color, size, languages etc. The detected text blocks in an image are represented by their bounding boxes. We evaluate the performance at the block level which is a common granularity level presented in the literature, [21, 28] rather than the word or character level. To judge the correctness of the text blocks detected, we manually count the true text blocks present in the frame. Further, we manually label each of the detected blocks as one of the following categories.

The truly detected text block (TDB) is a detected block that contains partially or fully text. The falsely detected text block (FDB) is a block with false detections. The text block with missing data (MDB) is a detected text block that misses some characters. Based on the number of blocks in each of the categories mentioned above, the following metrics are calculated to evaluate the performance of the method.

Detection Rate (DR) = Number of TDB / Actual number of text blocks
False Positive Rate (FPR) = Number of FDB / Number of (TDB + FDB)
Misdetection Rate (MDR) = Number of MDB / Number of TDB

The performance of the proposed method in comparison with the existing methods for the images taken from Horizontal-1 dataset is highlighted in Table I where we can see the detection rate, false positive rate and misdetection rate.

Table-I. Evaluation performance on Horizontal-1 dataset

| Methods | DR | FPR | MDR |
|---|---|---|---|
| Uniform Color based[34 ] | 51.3 | 27.3 | 37.3 |
| Gradient based[ 5] | 71.1 | 12.1 | 10.0 |
| Edge based[3 ] | 80.0 | 18.3 | 20.1 |
| Gradient + Edge based[21 ] | 88.54 | 4.72 | 2.56 |
| **Proposed method** | **92.45** | **3.85** | **6.32** |

By looking at Table-I, it shall be observed that the proposed method outperforms the other existing methods in terms of Detection Rate (DR) and Misdetection Rate (MDR). The DR of the proposed method is 92.45% and is higher than that of the existing methods. Similarly, the MDR of the proposed method is 6.32% which is lower than some of the existing methods. Therefore, the proposed method has achieved better detection results than the other four existing methods.

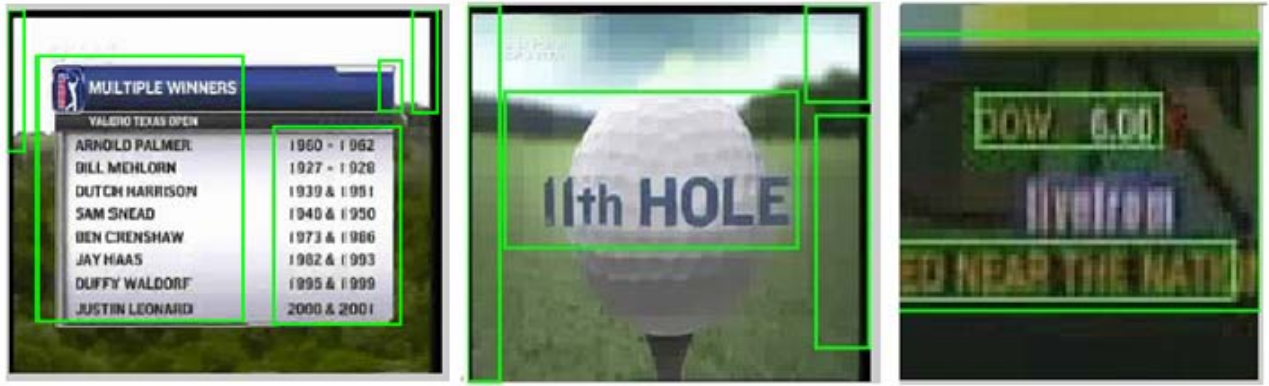

**a) Proposed method**

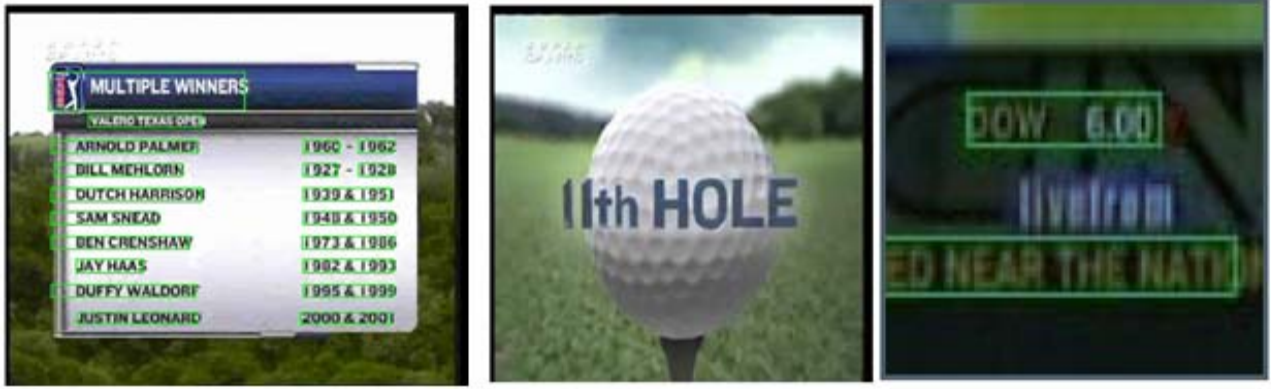

**b) Gradient + Edge based method**

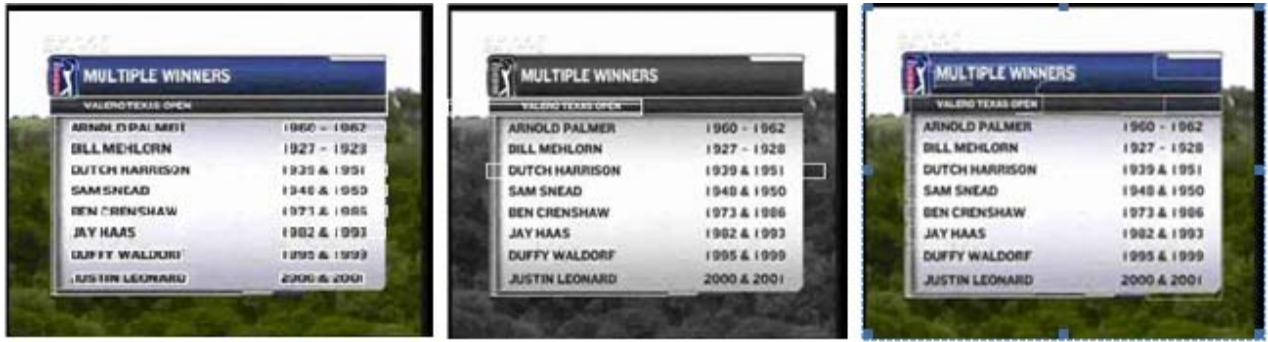

**c) Edge based method**     **d) Gradient based method**     **e) Uniform Color based method**

**Fig. 4.1 Sample Text Localization results for the images taken from Horizontal-1 dataset**

We have also conducted experimentation on some scene images where text is having poor illumination problem, low contrastness, blurring etc. The proposed method can detect text correctly for complex background image. It shall be noticed from Fig. 4.1(a) that the proposed approach is successful in localizing the text when it is blurred /low contrast but produces false positives. The edge based method also detects text similar to our proposed method. However, the gradient based method and uniform color based methods misses some text and they tend to include additional non text information in the bounding boxes. The edge based method fixes a single bounding box for two to three text lines. This is because of the threshold values fixed for classifying text pixel and non text pixel. Therefore, the detection rate of the method is quite low compared to the proposed method.

The gradient based method detects text with inaccurate boundaries as it works based on several constant thresholds for potential text line detection. The uniform color method also fails to detect text in the image as it fixes inaccurate boundaries for the text lines because of its limitation to handle illumination problem. It is observed from Fig. 4.1 (a) that the proposed method detects text even in the complex background image including scene text. On the other hand, existing methods fail to detect scene text in the image accurately as shown in Fig. 4.1(b). The detection rate, false positive rate and misdetection rate are therefore found be higher than the existing methods.

Experimentation was also performed on the images obtained from the publicly available bench mark dataset of ICDAR-2003 and the sample text localization results are shown in Fig. 4.2. These results show that the proposed method is capable of localizing the text but it produces few false alarms due to the reflectance of the material or due to the low contrast between the text and the background which needs to be addressed in future.

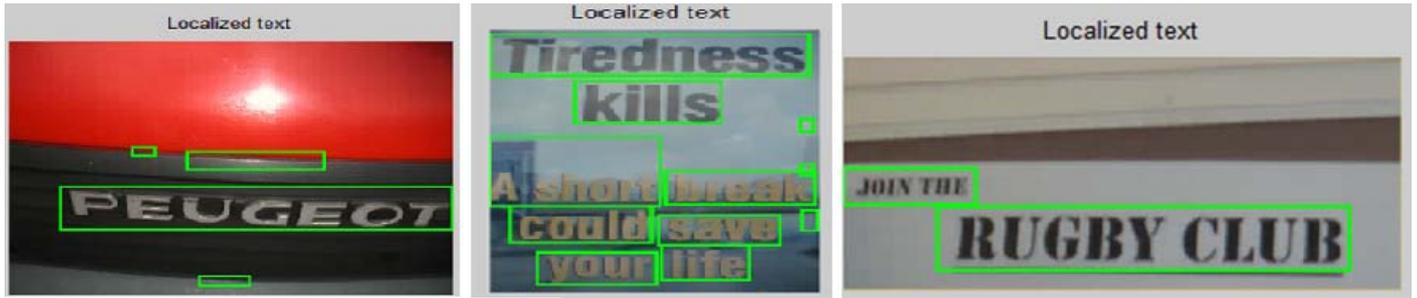

**Fig. 4.2 Sample Text Localization results from ICDAR-2003 dataset**

The Fig. 4.3 shows that the proposed method detects and localizes the scene text and caption text successfully but produces some false detections. The gradient difference approach helps in finding the pixels with strong intensity values that further localizing the text better. The sample results show that the proposed method is capable of detecting the small fonts with low contrast too. The method sometimes fails in eliminating few false negatives as shown in the Fig. 4.3 which is also localized along with the detected text. We have conducted experimentation on Hua dataset and observed a detection rate of 89.3%, false positive rate of 25.3% and misdetection rate of 5.7%.

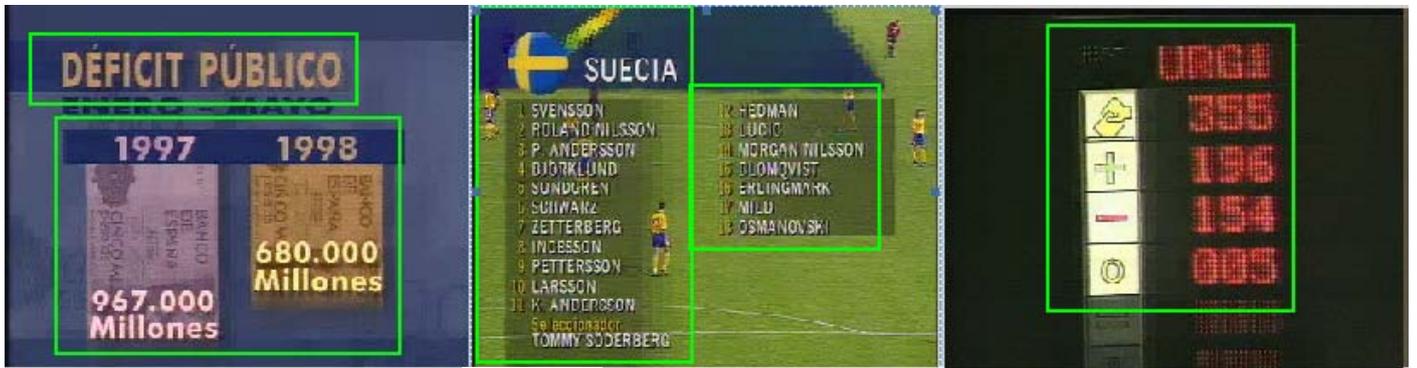

**Fig. 4.3 Sample Text Localization results from Hua dataset**

We have also conducted experimentation on some scene images having variation in font style, type, poor illumination and the results are shown in Fig. 4.4.

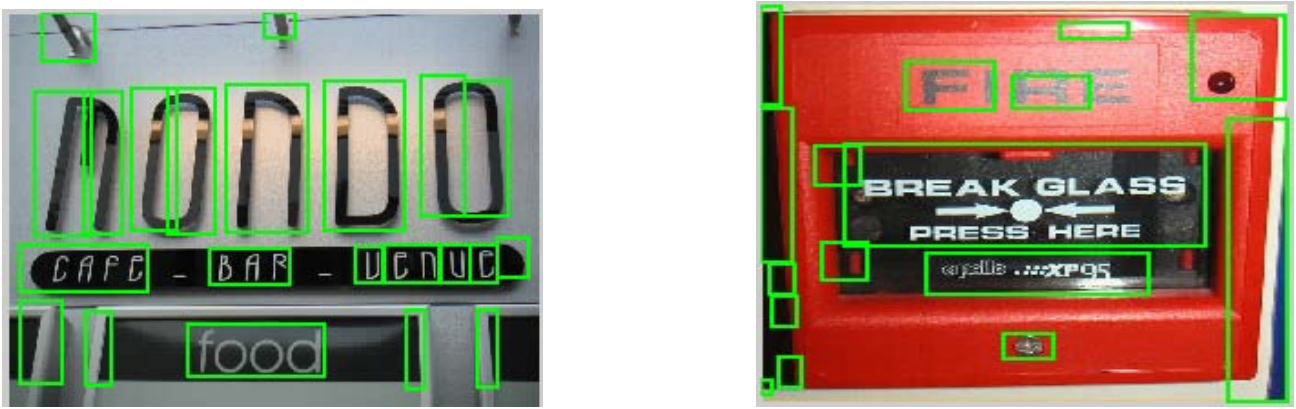

**Fig. 4.4 Sample Text Localization results for the images taken from ICDAR- 2003 dataset**

Some of the multi-oriented text blocks are shown in Fig. 4.5 where the first two blocks are partially detected and the third one (the bottom line is not detected in Fig.4.5a) whereas the proposed method detects all the three lines and localizes it. It shall be here observed that the image is of low contrast with multi-oriented text. The proposed method finds the text regions but fails to localize the lines separately.

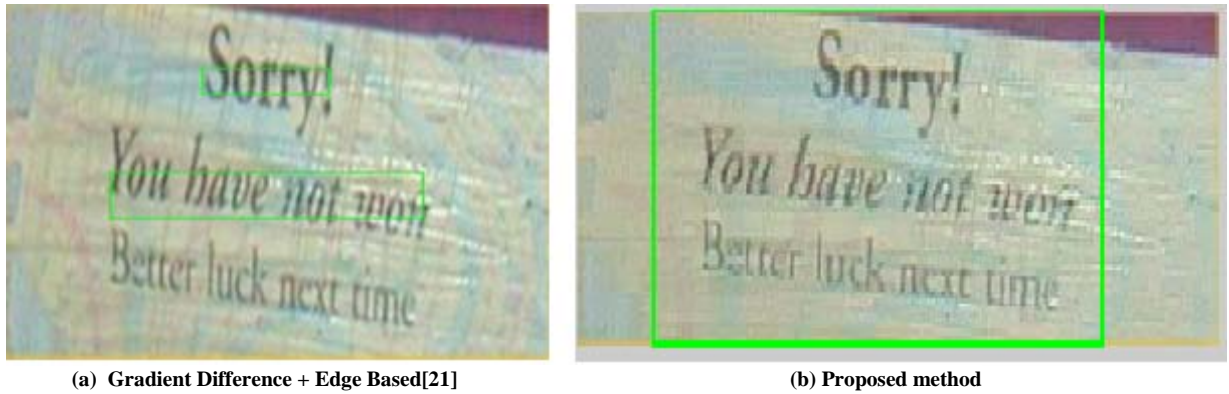

(a) **Gradient Difference + Edge Based[21]**        (b) **Proposed method**

**Fig. 4.5 Multi-Oriented Text Blocks that is partially detected/ not detected**

## 5. Conclusion

Text embedded in scene images/ video frames contain abundant high level semantic information which is important to analysis, indexing and retrieval. We developed a gradient difference based approach that helps in text localization under compressed domain. The newly developed approach is capable of localizing the text regions in scene images and video frames. Further, the method can be extended to fix the bounding boxes for the text lines separately. Experimental results show that the proposed method is capable of localizing the text but various problems that occur due to complex background need to be addressed in our future works.